%% file: main.tex
\documentclass[sigconf]{acmart}

\setcopyright{rightsretained}
\makeatletter
\gdef\@acmISBN{}
\gdef\@acmDOI{}
\makeatother

\AtBeginDocument{
  \fancyhead[LE]{\sffamily\footnotesize }
  \fancyhead[RO]{\sffamily\footnotesize }
  \fancyhead[LO]{\sffamily\footnotesize State-Inference-Based Prompting for Natural Language Trading with Game NPCs}
  }
\makeatother

\usepackage{natbib}
\setcitestyle{numbers,square,sort&compress}
\usepackage{tcolorbox}
\usepackage{listings}
\usepackage{xcolor}
\usepackage{subcaption}
\usepackage{kotex}
\usepackage{booktabs}
\usepackage{multirow}
\usepackage{makecell}
\usepackage{tikz}
\usetikzlibrary{positioning,arrows.meta,shapes}

\usepackage{enumitem}

\usepackage{xspace}
\usepackage{url}

\newcommand{\codeurl}[1]{\url{#1}}

\tcbuselibrary{listingsutf8}
\tcbuselibrary{breakable}
\newtcolorbox{promptbox}[1][]{%
  colback=gray!5!white, colframe=gray!80!black,
  fonttitle=\bfseries, title=Prompt Example,
  listing only, breakable,
  listing options={
    basicstyle=\ttfamily\small,
    breaklines=true,
    tabsize=2,
    keywordstyle=\bfseries\color{blue},
    commentstyle=\itshape\color{gray},
    stringstyle=\color{teal},
    numbers=left,
    numberstyle=\tiny\color{gray},
    frame=single,
    showstringspaces=false,
    columns=fixed,
    linewidth=\textwidth,
    literate={~}{{\textasciitilde}}1
  },
  #1
}
\definecolor{steelblue}{rgb}{0.27, 0.51, 0.7}
\definecolor{azure}{rgb}{0.0, 0.5, 1.0}
\definecolor{nicegreen}{rgb}{0.0, 0.7, 0.1}
\definecolor{CuGray}{gray}{0.9}
\definecolor{amethyst}{rgb}{0.6, 0.4, 0.8}
\definecolor{black}{rgb}{0.0, 0.0, 0.0}

\newcommand{\mk}[1]{\textcolor{black}{#1}}

\begin{document}

\title{State-Inference-Based Prompting \\for Natural Language Trading with Game NPCs}

\author{Minkyung Kim}
\affiliation{%
  \institution{SayBerryGames}
  \city{Daejeon}
  \country{South Korea}}
\email{mkkim1778@sayberrygames.com}

\author{Junsik Kim}
\affiliation{%
  \institution{Amazon}
  \city{Seattle}
  \country{U.S.A.}}
\email{jskimcv@amazon.com}

\author{Hwidong Bae}
\affiliation{%
  \institution{xl8}
  \city{Seoul}
  \country{South Korea}}
\email{bae.hwidong@gmail.com}

\author{Woongcheol Yang}
\affiliation{%
  \institution{SayBerryGames}
  \city{Daejeon}
  \country{South Korea}}
\email{woongy@sayberrygames.com}

\author{Sangdon Park}
\authornote{Co-corresponding authors.}
\affiliation{%
  \institution{SayBerryGames}
  \city{Daejeon}
  \country{South Korea}}
\email{chaos@sayberrygames.com}

\author{Sohee Bae}
\authornotemark[1]
\affiliation{%
  \institution{SayBerryGames}
  \city{Daejeon}
  \country{South Korea}}
\email{bread@sayberrygames.com}

\begin{abstract}
Large Language Models enable dynamic game interactions but struggle with rule-governed trading systems. Current implementations suffer from \mk{rule violations, such as item hallucinations and calculation errors, that erode player trust.} Here, State-Inference-Based Prompting (SIBP) enables reliable trading through autonomous dialogue state inference and context-specific rule adherence. The approach decomposes trading into six states within a unified prompt framework, implementing context-aware item referencing and place\-holder-based price calculations. Evaluation across 100 trading dialogues demonstrates \mk{>97\%} state compliance, \mk{>95\%} referencing accuracy, and 99.7\% calculation precision. SIBP maintains computational efficiency while outperforming baseline approaches, establishing a practical foundation for trustworthy NPC interactions in commercial games.
\end{abstract}




\settopmatter{printacmref=false}  
\maketitle

\input{sections/01_Introduction}
\input{sections/02_RelatedWork}
\input{sections/03_ProposedSystem}
\input{sections/04_ExperimentsResults}
\input{sections/05_Discussion}

\section{Conclusion}
This work introduces State-Inference-Based Prompting (SIBP), a comprehensive methodology that enables reliable LLM-driven trading through autonomous dialogue state inference and systematic rule adherence. By decomposing trading interactions into five distinct states within a unified prompt framework, SIBP addresses critical challenges that have hindered the deployment of LLM-based NPCs in commercial games: item hallucinations, inventory inconsistencies, and calculation errors that erode player trust.

Empirical evaluation demonstrates SIBP's effectiveness across multiple dimensions. The methodology achieves >97\% state transition compliance, >95\% item referencing accuracy, and 99.7\% price calculation precision, eliminating major barriers to commercial deployment. These results establish SIBP as a practical solution that bridges the gap between LLM capabilities and game system requirements while maintaining computational efficiency.

The broader implications extend beyond gaming. SIBP's structured approach to prompt engineering provides fundamental advances for rule-governed interactive systems where reliability and consistency are paramount. This methodology demonstrates how sophisticated state management can transform LLM capabilities into trustworthy, task-oriented systems, establishing a foundation for the next generation of intelligent interactive agents that combine natural language flexibility with structured system reliability.

\bibliography{sample-base}

\newpage
\appendix
\onecolumn

\input{sections/Appendix_NPCPrompts}

\input{sections/Appendix_UserPrompts}

\input{sections/Appendix_UserPrompts2}

\end{document}

%% file: sections/01_Introduction.tex
\section{Introduction}

Large Language Models (LLMs) are revolutionizing non-player character (NPC) interactions in games, enabling natural language-driven experiences that transcend traditional script constraints \citep{Luo2024LLMAgentsSurvey, Maleki2024LLMPCGSurvey}. These capabilities have spurred research and development of dynamic NPCs that enhance player immersion across diverse gaming contexts \citep{park2023generative, Song2025LLMDrivenNPCs, Li2025CharacterGen, Todorova2025QuestInfo, Gero2024PANGeA, Wang2025CivAgent}. Commercial implementations already demonstrate this potential in titles such as Relu Games' `Magical Mic Duel: Senpai, Hear My Spell', `Uncover the Smoking Gun' and Krafton's `inZOI' \citep{ReluGames2025MagicalMike, ReluGames2025SmokingGun, Krafton2025InZoiGDC}.

Trading mechanics represent a critical test case for LLM-driven interactions, integrating economic systems, social dynamics, and narrative progression under structured rule systems. Role-playing games particularly depend on merchant interactions as core experiences that bridge exploration, combat, and character development. Yet most games implement trading through static inventories, fixed pricing, and limited dialogue options \citep{kim2025mart}—constraints that prevent dynamic commercial interactions and limit emergent gameplay potential.

Recent research explores LLM-powered merchant NPCs but typically addresses isolated aspects rather than comprehensive interaction flows. Critical issues persist: item hallucinations, inventory inconsistencies, and rule violations that systematically undermine system reliability \citep{kim2025mart, Todorova2025QuestInfo}. Current LLM-based NPC research emphasizes social interactions \citep{park2023generative, Song2025LLMDrivenNPCs}, leaving rule-governed trading as a fundamental challenge \citep{Yuan2024RPGBench}.

This work develops a comprehensive framework that balances natural language flexibility with systematic rule adherence for robust trading systems. The framework enables LLMs to infer \mk{dialogue} states autonomously from the dialogue context without external state management. \mk{This is achieved by} decomposing trading into six distinct states with comprehensive guidelines within a unified prompt architecture. 

The proposed methodology leverages LLMs' generative capabilities while implementing robust control mechanisms to address hallucinations and ensure game system integrity. 
The approach enhances trading with three core innovations: \mk{structured state progression rules preventing accidental transactions}, dynamic item reference adjustment based on dialogue context, and placeholder-based post-processing ensuring mathematical accuracy without computational overhead.

Empirical evaluation \mk{conducted in} an operational role-playing game environment demonstrates the methodology's effectiveness through comprehensive metrics, including rule compliance rates, item reference accuracy, and price calculation precision. \mk{The results demonstrate} practical viability in commercial game environments.

\mk{This work makes the following key contributions:}
\begin{itemize}
    \item This study proposes a State-Inference-Based Prompting \mk{(SIBP)} methodology that enables autonomous dialogue state inference and rule compliance through four key \mk{prompt} design elements.
    \item The methodology decomposes trading into six states, ensuring procedural integrity while preventing unintended behaviors.
    \item The evaluation provides empirical validation through quantitative assessment demonstrating \mk{over 97\%} state transition compliance and \mk{deployment potential}.
    \item The \mk{SIBP methodology facilitates} context-aware data referencing mechanisms that achieve \mk{over 95\%} accuracy through dynamic adjustment while preventing hallucinations.
    \item The \mk{SIBP methodology enables} state-specific post-processing that ensures 99.7\% price calculation accuracy through place\-holder-based mechanisms.
\end{itemize}

The system architecture and prompt design principles offer broad generalizability for LLM-based interactive systems beyond gaming contexts, contributing fundamental advances to prompt engineering for rule-governed interactions. The implications extend to interactive kiosks, customer support systems, and educational tools, demonstrating how sophisticated interaction design can transform LLM capabilities into reliable task-oriented systems suitable for commercial deployment.

The remainder of this paper is organized as follows: Section 2 reviews related work; Section 3 details the proposed methodology; Section 4 provides experimental evaluation; Section 5 discusses significance and implications; Section 6 concludes with contributions and future directions.

%% file: sections/02_RelatedWork.tex
\section{Related Work}

LLMs have transformed interactive entertainment, creating new opportunities for dynamic player experiences \cite{Luo2024LLMAgentsSurvey, Maleki2024LLMPCGSurvey}. Traditional NPCs operate through pre-written dialogues and finite state machines, resulting in predictable interactions that reduce immersion. Recent research explores LLM-driven approaches for naturalistic and adaptive NPC dialogues, demonstrating how language models can drive agents with sophisticated social interactions and persistent memory capabilities.

Park et al. pioneered Generative Agents, demonstrating that LLMs can generate believable human-like behavior through integrated long-term memory, reflection, and planning mechanisms \cite{park2023generative}. Marincioni et al. investigated how emotional states expressed by LLM-based NPCs affect player responses using emotion extraction methods \cite{marincioni2024emotions}. Recent studies have also examined NPC interactions in VR environments, comparing different communication methods with LLM-based characters \cite{christiansen2024presence}. However, these studies focus primarily on open-ended conversations rather than structured, goal-oriented interactions requiring strict rule adherence.

Several studies have examined commercial interactions and negotiation dynamics using LLMs. Abdelnabi et al. proposed a benchmark for evaluating LLMs in multi-agent negotiation scenarios \cite{abdelnabi2024llmstakeholders}. Chatterjee et al. introduced \codeurl{AgreeMate}, a framework for teaching LLMs strategic price haggling \cite{Chatterjee2024AgreeMate}, focusing primarily on bargaining aspects rather than broader procedural requirements. Kim et al. developed the \codeurl{MART} framework for active merchant NPCs capable of dynamic pricing \cite{kim2025mart}. While \codeurl{MART} achieved progress in natural language trading, persistent challenges remain in \mk{maintaining consistent protocol adherence,} preventing item hallucination, \mk{and} ensuring accurate calculations.

Previous research has contributed valuable innovations across multiple dimensions of interactive system design. However, existing approaches typically focus on specific functionalities such as price negotiation \cite{kim2025mart} or emotion expression \cite{marincioni2024emotions}, rather than addressing comprehensive requirements of dynamic trading dialogues involving multiple rule-based states including inventory browsing, item selection, and multi-step confirmation procedures.

Such comprehensive trading systems require strict adherence to inventory constraints, pricing rules, and procedural protocols throughout entire dialogue flows while maintaining natural conversational patterns. A significant research gap exists in prompt engineering methodologies that enable LLMs to infer current trading states from dialogue context alone—essential for reliable dynamic trading interactions that maintain compliance with complex game system rules. \mk{While Dialogue State Tracking research \cite{Feng2023DSTInstruction} is crucial for this state inference, it often does not extend to controlling rule-governed dialogue flow.}

The State-Inference-Based Prompting methodology introduced in this study systematically addresses these limitations through a unified approach to structured dialogue management. This methodology enables LLMs to autonomously infer trading states from dialogue context and maintain adherence to game system rules throughout complete trading processes without requiring external state management systems.

%% file: sections/03_ProposedSystem.tex
\section{Proposed System}

This section details the system architecture and core prompting methodology developed for natural language trading with game NPCs. The proposed State-Inference-Based Prompting approach guides LLMs to follow predefined dialogue states within goal-oriented interactions, with in-game trading serving as the primary application domain.

\subsection{LLM-Driven Interaction Framework}
\label{sec:InteractionFramework}

The system enables players to naturally converse and trade with NPC merchants through a multi-stage process: (1) collect player input, (2) combine with game information including NPC inventory and dialogue history to construct contextual prompts, (3) call LLM API, (4) parse responses and update the game state, (5) present NPC dialogue to the player.

The most challenging aspect is the fundamental \textbf{unpredictability of dialogue}. Unlike menu-based trading in traditional games, natural language conversations allow players to change context dynamically at any time. Representative examples include:

\begin{itemize}
    \item While an NPC is explaining item lists, suddenly asking \emph{"How much is this?"} (immediate price inquiry)
    \item After getting price information, immediately proposing \emph{"Can you make it 50 gold?"} (direct negotiation attempt)
    \item During active trade negotiations, suddenly asking \emph{"Are there many monsters around lately?"} (casual conversation)
    \item Even abrupt \emph{"Goodbye"} statements (conversation termination)
\end{itemize}

Since predicting player utterances is impossible, NPCs must \textbf{judge the current situation autonomously after listening to player input}. However, if the LLM misjudges the situation, inappropriate responses may occur, such as suddenly proceeding to payment during casual chat or failing to maintain transaction state consistency.

To achieve this capability, the prompt is structured to enable LLMs to accurately judge situations and respond according to game rules. Key components include:

\begin{itemize}
    \item \textbf{System Instructions}: Guidelines defining the NPC's persona, role, and behavioral directives.
    \item \textbf{Game World Data}: Lists of all items available in the game and the NPC's inventory with item ID, name, quantity, and price.
    \item \textbf{Situational Context}: Information about the NPC's \mk{profile} and current in-game environment such as current location and time.
    \item \textbf{Dialogue History}: A record of preceding dialogue turns between player and NPC, including the latest player input and inferred context from previous NPC turns.
    \item \textbf{Contextual and Trade Guidelines}: Rules and guidelines to infer states and identify when to transition between states (\codeurl{<CONTEXT_GUIDELINES>} and\\ \codeurl{<TRADE_GUIDELINES>}\mk{; the full prompt is provided in Appendix \ref{sec:npc_prompt}}).
    \item \textbf{Expected Response Format}: Instructions for the LLM to structure its response in a format that facilitates system integration.
\end{itemize}

As guided by the Expected Response Format instructions, the LLM generates output as a single, well-formed JSON object. This approach relies on the LLM's capability to adhere to formatting instructions provided through prompting, rather than utilizing specialized structured output functionalities. The resultant JSON response facilitates seamless parsing and integration with game systems.

Key fields in the response structure include:
\begin{itemize}
    \item \textbf{\codeurl{context_reason}} (string): A summary or reasoning behind the LLM's understanding of the current context.
    \item \textbf{\codeurl{context_type}} (string): The conversational context inferred by the LLM.
    \item \textbf{\codeurl{context_details}} (object): Contains finer-grained context information, particularly for \codeurl{TRADE} context. This includes the trade subcontext and an array of items pertinent to the current trade interaction.
    \item \textbf{\codeurl{npc_dialogue}} (string): The natural language dialogue spoken by the NPC to the player.
\end{itemize}

This framework ensures that LLMs receive rich contextual information through detailed prompts and provide structured, actionable responses, enabling goal-oriented and rule-consistent NPC interactions that maintain both game world integrity and conversational naturalness.

\subsection{State-Inference-Based Prompting}

To address the dialogue unpredictability problem mentioned above, this work proposes \textbf{State-Inference-Based Prompting (SIBP)}. The core innovation of SIBP is enabling LLMs to determine \emph{"what situation is happening right now?"} autonomously from the dialogue context alone.

The system defines three main situational contexts:
\begin{itemize}
    \item \textbf{\codeurl{NONE}} (general conversation): Casual interactions such as \emph{"Hello,"} \emph{"Nice weather,"} \emph{"Monster news,"} etc.
    \item \textbf{\codeurl{TRADE}} (trading conversation): Commercial interactions such as \emph{"I want to buy a sword,"} \emph{"How much?"} \emph{"Can you lower the price?"} etc.
    \item \textbf{\codeurl{END_CONVERSATION}} (conversation ending): Termination phrases such as \emph{"Goodbye,"} \emph{"I'll come back later,"} etc.
\end{itemize}

The LLM learns the characteristics of each situation from the \codeurl{<CONTEXT_GUIDELINES>} section in the prompt, judges the appropriate situation based on player utterances, and then responds following rules appropriate for that situation.

For example, if a player suddenly asks \emph{"Are there many monsters around lately?"} the LLM judges this as a \codeurl{NONE} situation and responds with monster-related casual conversation instead of trade-related responses. This solves the \emph{"sudden context changes"} problem mentioned in Section 3.1.

\input{objects/fig_teaser}

\subsubsection{Trade Subcontexts}

However, simply having one \codeurl{TRADE} state is insufficient for implementing proper trading systems. For example, when a player says \emph{"I want to buy a sword,"} \emph{"How much is that sword?"} and \emph{"Yes, I'll buy it,"} the NPC's responses should be completely different for each case, requiring distinct behavioral guidelines.

If all these situations were handled with just a simple \codeurl{TRADE} state:
\begin{itemize}
    \item \emph{"I want to buy a sword"} → Unclear whether NPC should immediately state the price or ask which sword
    \item \emph{"How much?"} → Cannot determine which item's price is being asked or if negotiation is possible
    \item \emph{"Yes, I'll buy it"} → It is unknown whether to proceed directly to payment or require additional confirmation
\end{itemize}

Therefore, the trading process must be decomposed more granularly to clearly define how NPCs should behave at each stage. For the scope of this study, focusing on player buying from an NPC, the following \mk{five} trade subcontexts (i.e., sub-states) are defined:

\begin{itemize}
    \item \textbf{\codeurl{SHOW_INVENTORY}}: The state where the NPC displays or mentions items available for sale to the player.
    \item \textbf{\codeurl{OFFER_SELL}}: The state where the NPC proposes selling specific items and states their price.
    \item \textbf{\codeurl{NEGOTIATE_PRICE}}: The state where the player attempts to negotiate the price, and the NPC responds based on their character traits.
    \item \textbf{\codeurl{CHECK_CONFIRMATION}}: A state where the NPC seeks final confirmation from the player before finalizing a purchase, designed to prevent abrupt transaction completion.
    \item \textbf{\codeurl{CONFIRM_SELL}}: The state where the player confirms their purchase intent, leading to the final update of game data (e.g., deducting currency, removing items from inventory).
\end{itemize}

These states are not strictly sequential and can transition dynamically based on player input. For instance, a player might return to requesting other items while confirming a purchase, or switch between general conversation (\codeurl{NONE}) and trade interaction (\codeurl{TRADE}) to discuss the rarity or demand of specific items. This dynamic behavior requires an LLM approach capable of robust state inference.

\subsubsection{Prompt Design Elements}

But how can the system make LLMs properly distinguish and manage multiple trading states? Simply instructing \emph{"respond appropriately to the situation"} leads to problems:

\begin{itemize}
    \item \textbf{State confusion}: When a player says \emph{"Yes, that sounds good,"} the LLM cannot distinguish whether this is accepting a price offer or simply a positive reaction to an item description
    \item \textbf{State skipping}: Players saying \emph{"I'll buy it"} immediately jumps to payment, skipping final confirmation steps
    \item \textbf{Inaccurate state inference}: LLMs failing to correctly identify the previous trading state from dialogue context alone, leading to inappropriate state transitions
    \item \textbf{Unclear reasoning}: Unable to understand why the LLM made certain state judgments, making debugging and improvement difficult
\end{itemize}

To solve these problems, SIBP incorporates four key design elements into the prompt within the \codeurl{<TRADE_GUIDELINES>} sections. These elements are: (1) Basic State Definitions, (2) State Transition Conditions with Previous Context, (3) Directive to Identify Previous State, and (4) Directive to Respond including Previous State.

\paragraph{\textbf{Element 1: Basic State Definitions}}
Provides foundational definitions for each trading state by establishing two aspects. First, it establishes the \textbf{context} of each state—what situation each state represents. For example, \codeurl{SHOW_INVENTORY} is defined as \emph{"the state where NPCs present available items to players,"} while \codeurl{OFFER_SELL} is characterized as \emph{"the state where NPCs propose specific items with pricing."} Second, it specifies the \textbf{behavioral rules} for each state—what actions should be taken within that state. This foundational definition ensures that LLMs understand both the situational context and appropriate behavioral guidelines for each state.

\paragraph{\textbf{Element 2: State Transition Conditions with Previous Context}}
Solves both the \textbf{state confusion} and \textbf{state skipping} problems by defining the prerequisite conditions for entering each trading state, emphasizing the dependency on previous states. This element addresses state confusion by providing context-aware transition rules—for example, when a player says \emph{"Yes, that sounds good,"} the appropriate response depends on whether the previous state involved item description (\codeurl{SHOW_INVENTORY}) or price offering (\codeurl{OFFER_SELL}). It prevents state skipping by establishing mandatory sequential flows. For instance, \codeurl{CONFIRM_SELL} is accessible only when the previous state was \codeurl{CHECK_CONFIRMATION} and the player provides positive confirmation.

\paragraph{\textbf{Element 3: Directive to Identify Previous State}}
Addresses the \textbf{inaccurate state inference} problem by instructing the LLM to identify the previous trading state before determining the current response. The prompt includes commands such as \emph{"Before responding, identify what the last trading state was based on the dialogue history."} This directive addresses the core challenge where LLMs may understand general conversation context but fail to accurately recognize the specific state identifiers from previous interactions.

\paragraph{\textbf{Element 4: Directive to Respond including Previous State}}
Addresses both the \textbf{inaccurate state inference} problem and the \textbf{unclear reasoning} problem by requiring the LLM to document its \mk{identification of the previous trading state}. While Element 3 focuses on INPUT processing (\emph{"identify the previous state"}), Element 4 mandates OUTPUT transparency (\emph{"document your state identification"}). The JSON response must include a \codeurl{last_trade_context} field where the LLM records the identified previous state. This documentation enables system monitoring of the LLM's reasoning process, facilitates debugging, and provides a foundation for the improvement of state inference accuracy.

\vspace{5mm}

Figure \ref{fig:teaser} illustrates the core concept of SIBP through a concrete example. The figure demonstrates how the same player utterance \emph{"That sounds good!"} triggers completely different NPC behaviors depending on the previous dialogue state. In the upper scenario, when the previous state was \codeurl{SHOW_INVENTORY}, the LLM interprets the player's response as interest in an item and transitions to \codeurl{OFFER_SELL}, providing a price proposal. In the lower scenario, when the previous state was \codeurl{OFFER_SELL}, the same utterance is interpreted as price acceptance and transitions to \codeurl{CHECK_CONFIRMATION}, seeking purchase confirmation. This context-aware state inference enables NPCs to maintain coherent dialogue flows while autonomously managing complex trading interactions without external state management systems.

\subsection{Additional State-specific Rules}

The expansion of SIBP by adding state-specific guidelines yields advantages in implementing game system functionalities and ensuring rule adherence. The system can achieve the following capabilities:

\subsubsection{State-specific Data Referencing}

This feature dynamically adjusts the data that NPCs reference based on dialogue context, addressing a challenge in maintaining both game world knowledge and transaction accuracy.

\textbf{How it works}: By defining item reference conditions based on states, the LLM responds differently:
\begin{itemize}
    \item \textbf{\codeurl{NONE} state (general conversation)}: References the complete game item list (\codeurl{<GAME_ITEMS_LIST>}) for general world discussions
    \item \textbf{\codeurl{TRADE} state (trading conversation)}: References only the NPC's inventory (\codeurl{<CHARACTER_INVENTORY>}) for actual sellable items
\end{itemize}

\textbf{Benefits}: This enhances player immersion while ensuring accurate transaction capabilities. NPCs can naturally discuss various game world items during the general conversation but only propose actually sellable items during trading, preventing item hallucinations while maintaining conversational richness.

\subsubsection{State-specific Post-processing}

As highlighted in prior research \citep{kim2025mart}, LLMs, being language models, can face challenges in precise mathematical computations. Even when correctly listing item quantities and individual prices, errors may occur in final price calculations, compromising transaction integrity.

\textbf{Solution}: SIBP implements accurate price calculations through state-specific post-processing. In the \codeurl{OFFER_SELL} state—the initial price offering stage where price negotiation is not yet required—the system works as follows:

\begin{enumerate}
    \item \textbf{Placeholder usage}: The prompt includes the rule \emph{"use \codeurl{__PRICE__} for the final price amount"}
    \item \textbf{Accurate calculation}: The system calculates the correct price based on responded item information and replaces the placeholder
    \item \textbf{Error prevention}: The accurately calculated price is included in dialogue history, ensuring correct price references in subsequent conversations
\end{enumerate}

\textbf{Response example}:
\begin{itemize}
    \item \textbf{Original LLM response}: \emph{"Two iron swords and an adventurer's sleeping bag are \codeurl{__PRICE__} gold together."}
    \item \textbf{After system post-processing}: \emph{"Two iron swords and an adventurer's sleeping bag are 150 gold together."}
\end{itemize}

\textbf{Advantages}: This approach achieves 99.7\% accurate price calculation without additional computational overhead, eliminating the need for complex schemas or 
\mk{Tool use} mechanisms. The single response processing maintains system efficiency while ensuring mathematical precision. This state-based prompting approach demonstrates how external system functionalities such as accurate calculation and post-processing can be integrated into LLM-driven interactions, providing a foundation for broader applications in rule-governed interactive systems.

%% file: objects/fig_teaser.tex
\begin{figure*}[t]
    \centering
    \includegraphics[width=0.95\linewidth]{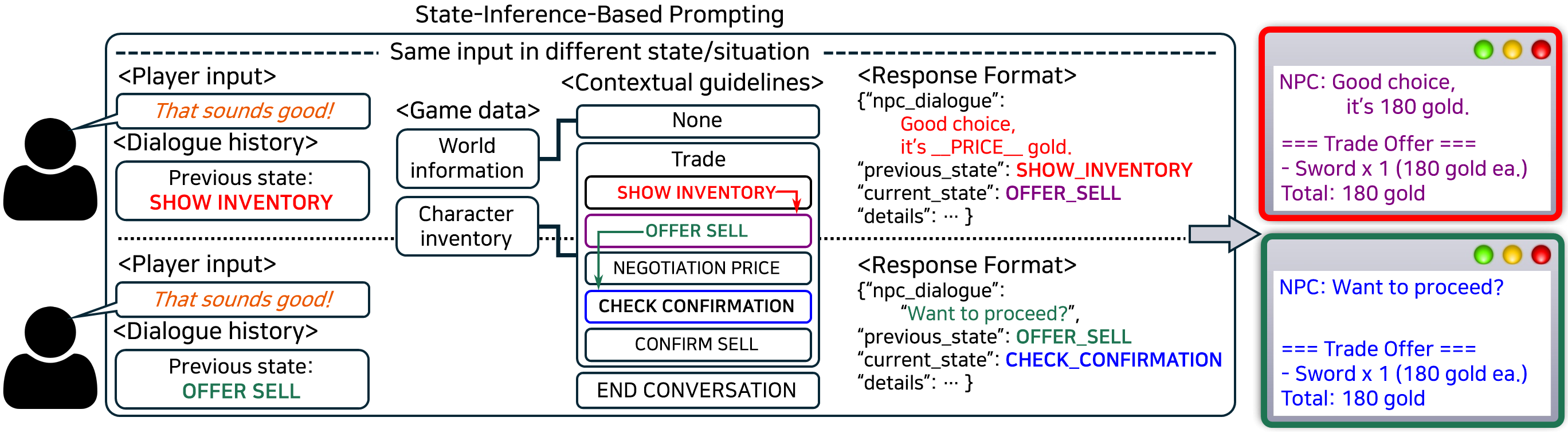}
    \caption{State-Inference-Based Prompting (SIBP) enables context-aware NPC responses. The same player utterance "That sounds good!" triggers different behaviors depending on the previous dialogue state: \codeurl{SHOW\_INVENTORY} → \codeurl{OFFER\_SELL} (price proposal) vs. \codeurl{OFFER\_SELL} → \codeurl{CHECK\_CONFIRMATION} (purchase confirmation). SIBP autonomously infers states from dialogue context while maintaining game system consistency.}
    \label{fig:teaser}
\end{figure*}

%% file: sections/04_ExperimentsResults.tex
\section{Experiments and Results}

This section presents comprehensive empirical evaluations of the State-Inference-Based Prompting (SIBP) methodology. The evaluation systematically assesses how effectively NPC LLMs adhere to predefined conversational flows during dynamic trading interactions and then demonstrates additional advantages of this approach through state-specific data referencing and post-processing.

\subsection{Experimental Setting}

All experiments employ conversations between a player-mimicking LLM (virtual player) and an NPC LLM to ensure controlled, reproducible evaluation conditions. One hundred independent trading dialogues were conducted using fixed seeds ranging from 0 to 99.
In each dialogue, the virtual player initiated conversations with the NPC LLM. While the NPC LLM's prompt varied across different experimental conditions as detailed in subsequent subsections, the initial utterance from the virtual player remained consistent for each seed, establishing controlled starting points for interactions.

\input{objects/tab_exp1}
\input{objects/fig_exp1}

The conversations were structured around two systematically designed test scenarios:
\begin{enumerate}
    \item \textbf{Request for Specific Item Purchase:} Virtual players initiate conversations by requesting specific items. Items are randomly selected from sellable and unsellable categories up to six in total. Sellable items refer to items with sufficient stock in the NPC's inventory \mk{(20 items)}, while unsellable items exist in the game world but are absent from the NPC's inventory \mk{(an additional 32 items, totaling 52 distinct items in the game)}. Purchase quantities for each item are randomly selected from 1 to 5. 
    For example: \emph{``I'd like to purchase 3 basic iron swords and 5 mana potions.''}
    \item \textbf{Request for Item Recommendation:} Virtual players initiate conversations mentioning purchase purposes such as \emph{``Could you recommend me some items for goblin battle?''}
\end{enumerate}

The virtual player prompts for each scenario are provided in Appendices \ref{sec:player_prompt} and \ref{sec:player_prompt2}. Virtual players and NPCs engage in alternating dialogue turns. Conversation length is defined as the number of interaction rounds, where each round consists of one player turn followed by one NPC turn. 
\mk{Average conversation lengths were 5.17 rounds (SD=1.0, range: 3-8) for Scenario 1 and 6.05 rounds (SD=1.6, range: 2-12) for Scenario 2.}

Unless specified otherwise in individual experiment descriptions, both virtual player LLMs and NPC LLMs utilized the \codeurl{gemini-2.5-flash-preview-04-17} model with a temperature setting of 0.7 and a thinking budget of 0.

\subsection{\mk{State Transition Compliance}}

A key advantage of SIBP lies in its ability to clearly define states and transitions in goal-oriented dialogues and systematically guide LLMs to follow them. SIBP achieves this through four core elements: (1) explanation of states, (2) explanation of state transition conditions, (3) directive to identify the last previous state, and (4) directive to include the last previous state in the response.

\textbf{Experimental Design}: To precisely measure the effect of each element, experiments systematically activated/deactivated the four elements to create multiple baselines. The complete SIBP includes all elements, while baselines activate only some elements. Table \ref{tab:state_transition_compliance} shows which elements are activated (\emph{`O'}) or deactivated (\emph{`X'}) for each method and their corresponding performance. This allows analysis of each element's individual contribution and the synergistic effects of their combinations.

\textbf{Evaluation Metric}: State Transition Compliance Rate (STCR) serves as the evaluation metric in this experiment. STCR is the proportion of conversations that correctly followed the \emph{``\codeurl{CHECK_CONFIRMATION} → \codeurl{CONFIRM_SELL}''} sequence. 
\mk{To ensure a fair comparison across methods with varying numbers of \codeurl{CONFIRM_SELL} state occurrences due to dialogue randomness, STCR was calculated based on the first 88 dialogues for each method where this state was reached. This count represents the minimum number of such occurrences observed across all methods within the 100 dialogues.}
Adherence to this specific state progression sequence was a critical rule intentionally enforced upon the LLM through the proposed prompt design. This sequence serves as an important procedural safeguard within game environments, designed to prevent unintended transactions by ensuring explicit player confirmation before purchase finalization.

\textbf{Key Results}: As shown in Table \ref{tab:state_transition_compliance}, SIBP with all elements achieved
\mk{a compliance rate of over 97\%}.
Element 4 
\mk{proved impactful, improving the rate to 94\% even without Element 3.}
\mk{Interestingly, while Element 3 did not by itself yield an improvement over simpler baselines (cf. baseline3 vs. baseline2), its combination with Element 4 was crucial for achieving the optimal compliance rate.}
This demonstrates that when LLMs identify previous states from dialogue history and explicitly include them in responses, adherence to designed dialogue flows is greatly enhanced.

\textbf{State Transition Analysis}: Figure \ref{fig:state_transitions} shows the observed state transitions as heatmaps. Each cell value matrix[i][j] represents the number of transitions from state i (previous) to state j (current). 
\mk{SS (Session Start) represents a conceptual point prior to the first dialogue turn. SE (Session End) signifies the conceptual termination point occurring after the final NPC response.}

Examining scenario-specific characteristics, Scenario 1 (direct specific item purchase) primarily begins with \codeurl{OFFER_SELL} (Figure \ref{fig:state_transitions_sc1}), while Scenario 2 (recommendation-based purchase) mainly starts with \codeurl{SHOW_INVENTORY} (Figure \ref{fig:state_transitions_sc2}). In both cases, valid state transitions were observed as designed, confirming that SIBP successfully induces natural dialogue flows appropriate to each situation.

\input{objects/tab_exp3}
\input{objects/fig_exp2-1}

\subsection{State-specific Data Referencing}

NPCs must balance natural conversation about the game world with accurate trading behavior. In general conversation, they can discuss all game items, but during trading, they should only propose items they actually sell, addressing the fundamental challenge of maintaining both world knowledge and transaction accuracy.

\textbf{Experimental Design}: This evaluation assesses SIBP's ability to reference different datasets based on dialogue state. \mk{The NPC} prompt includes two datasets: \codeurl{<GAME_ITEMS_LIST>} (IDs and names of all valid game items) and \codeurl{<CHARACTER_INVENTORY>} (IDs, names, quantities, and prices of NPC's sellable items). The challenge is making the LLM select the appropriate dataset based on context.

\textbf{Evaluation Metric}: Sellable Item Response Rate (SIRR) serves as the evaluation metric. 
\mk{In this system, all NPC responses generated within the \codeurl{TRADE} state are designed to populate an \codeurl{items} field with relevant item data. SIRR is then calculated as the overall proportion of these \codeurl{TRADE} state responses where the items field correctly contains only sellable item data. This calculation is aggregated across all 100 dialogues and provides a global measure of accuracy at the response level.}

\textbf{Key Results}:
\mk{The SIRR was 95.05\% in Scenario 1 and 95.88\% in Scenario 2 out of 485 and 534 NPC responses within the \codeurl{TRADE} state, respectively. 
The most frequent error type involved the NPC assigning a price of 0 or a null value for unsellable items. In rare responses (0.4\% and 0.2\% in Scenario 1 and 2, respectively), the NPC failed to provide any item information when expected within a trade context. Overall, these high SIRR values, coupled with a clear understanding of the failure cases, demonstrate SIBP's effectiveness in state-specific data referencing.}

\subsection{State-specific Post-processing}
\label{sec:exp_ppp}

LLMs struggle with precise arithmetic operations, making simple calculation errors like \emph{``2×50 + 30 = 120 gold.''} This poses significant problems for price calculations in trading, potentially undermining player trust and transaction integrity.

\textbf{Experimental Design}: 
\mk{Utilizing dialogues from Scenario 1,}
experiments compare 6 configurations 
(Table \ref{tab:price_accuracy}):
\begin{itemize}
\item \textbf{SIBP variants}: Basic SIBP, SIBP+PPP (with Placeholder Post-Processing), SIBP+SO (with Structured Output)
\item \textbf{Model variants}: 2.5-flash (baseline), 2.0-flash (lighter version), 2.5-pro (high-performance)
\end{itemize}

\textbf{Evaluation Metric}: The evaluation measures price accuracy,
completion tokens, thought tokens, and response time in \codeurl{OFFER_SELL} state and other price-related states.
\mk{Price accuracy is the percentage of NPC responses where the total price stated by the LLM matches the sum calculated from the item details (i.e., prices and quantities) also provided within the same LLM response.}

\textbf{Key Results}: 
SIBP+PPP showed consistent improvements in accuracy while maintaining computational efficiency across multiple evaluation dimensions.
First, in terms of accuracy, SIBP+PPP reached 100.0\% in the \codeurl{OFFER_SELL} state, compared to 79.7\% with basic SIBP. This improvement also extended to subsequent states, where SIBP+PPP achieved 99.7\% accuracy versus 82.3\% for basic SIBP. These results suggest that accurate initial pricing may help reduce downstream errors in multi-turn dialogues. 

Second, these accuracy gains were achieved without notable increases in computational cost. Token usage (398.6 vs 394.8) and response time (2.3 seconds) remained comparable to basic SIBP, while thought token usage was held at zero. By contrast, SIBP+SO introduced additional complexity, requiring 371.8 thought tokens and a longer response time of 4.1 seconds. Furthermore, the high variability in thought tokens observed in other configurations (e.g., 553.7 for SIBP+SO, 1013.6 for 2.5-pro) reflects the differing internal reasoning needs across dialogue types, with simpler exchanges requiring fewer resources and more complex negotiations involving significantly more.

Third, SIBP+PPP also showed promising results with smaller models. When tested with 2.0-flash, it achieved 100.0\% accuracy in \codeurl{OFFER_SELL} and 91.8\% in other states, outperforming the basic SIBP setup, which achieved 49.4\% and 67.3\%, respectively. This indicates the method’s potential for generalization across model variants.

Lastly, SIBP+PPP using 2.5-flash reached the same level of accuracy as Basic SIBP with the 2.5-pro (100.0\%) while requiring significantly fewer resources: 0 vs 2398.5 thought tokens and 2.3 vs 27.7 seconds in response time. These findings suggest that incorporating state-specific post-processing may allow smaller models to approximate the accuracy of larger ones, with considerably lower computational demands.

%% file: objects/tab_exp1.tex

\begin{table*}
    \centering
    \caption{State Transition Compliance Rate (STCR) for different prompt element combinations. STCR represents the proportion of conversations that correctly followed the mandatory confirmation step
    before finalizing transactions, serving as a safeguard to prevent unintended purchases. 
    In this table, the SIBP configuration also includes placeholder post-processing (PPP). Experiments in Section (\ref{sec:exp_ppp}) will differentiate SIBP (not using PPP) from SIBP+PPP.
    }
    \label{tab:state_transition_compliance}
    \begin{tabular}{cccccccc} 
        \toprule
        \multirow{2}{*}{\makecell{Scenario}} & 
        \multirow{2}{*}{\makecell{Method}} & \multicolumn{4}{c}{Prompt elements} & \multirow{2}{*}{\makecell{STCR [\%]}} \\
        \cline{3-6} 
        & & (1) Explain States & (2) Explain Transitions & (3) Identify Prev. State & (4) Response Prev. State & \\ 
        \midrule
        \multirow{5}{*}{\makecell{1}} 
        & baseline1       & O & X & X & X & 79.55 \\ 
        & baseline2       & O & O & X & X & 84.09 \\ 
        & baseline3       & O & O & O & X & 77.27 \\ 
        & baseline4       & O & O & X & O & 94.32 \\ 
        & SIBP            & O & O & O & O & 97.73 \\ 
        \midrule
        2 
        & SIBP & O & O & O & O & 97.73 \\ 
        \bottomrule
    \end{tabular}
\end{table*}

%% file: objects/fig_exp1.tex
\begin{figure*}
    \centering
    \begin{subfigure}{0.37\textwidth}
        \includegraphics[width=0.95\linewidth]{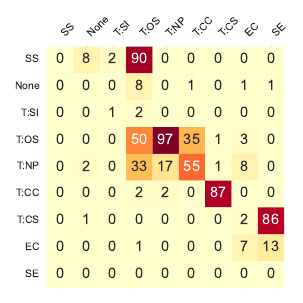}
        \caption{Scenario 1: the conversation initiated by requesting to purchase specific items}
        \label{fig:state_transitions_sc1}
    \end{subfigure}
    \hspace{0.05\textwidth}
    \begin{subfigure}{0.37\textwidth}
        \includegraphics[width=0.95\linewidth]{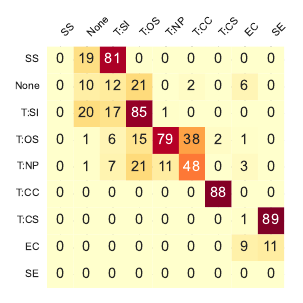}
        \caption{Scenario 2: the conversation initiated by asking for item recommendations based on a purchase purpose}
        \label{fig:state_transitions_sc2}
    \end{subfigure}
    \caption{State transition patterns in 100 dialogues. Cell values at row $i$ and column $j$ represent transitions from state $i$ to state $j$. Abbreviations; \mk{SS(Session Start)}, T:SI(Show Inventory), T:OS(Offer Sell), T:NP(Negotiate Price), T:CC(Check Confirmation), T:CS(Confirm Sell), \mk{SE(Session End)}.}
    \label{fig:state_transitions}
\end{figure*}

%% file: objects/tab_exp3.tex
\begin{table*}[t]
    \centering
    \caption{Performance comparison across six system configurations. Data is presented for the \codeurl{OFFER_SELL} state and for Others (a collective of \codeurl{NEGOTIATE_PRICE}, \codeurl{CHECK_CONFIRMATION}, and \codeurl{CONFIRM_SELL} states). \mk{The X/Y notation in each column header indicates the total number of NPC responses within the \codeurl{OFFER_SELL} state (X) and Others states (Y), respectively, over 100 dialogues.} Price accuracy is a percentage; Token usage and response time values are mean (standard deviation).
    }     
    \label{tab:price_accuracy}
    \vspace{-2mm}
    \begin{tabular}{ll|ccc|cc|c}
        \toprule
        \multirow{3}{*}{\makecell{\textbf{Metric} \\ }} & 
        \multirow{3}{*}{\makecell{\textbf{State} \\ }} &  
        \multirow{3}{*}{\makecell{\textbf{SIBP}  \\ \textbf{2.5-flash} \\ (192/283)}} & 
        \multirow{3}{*}{\makecell{\textbf{SIBP+SO} \\ \textbf{2.5-flash} \\ (209/294) }} &
        \multirow{3}{*}{\makecell{\textbf{SIBP+PPP} \\ \textbf{2.5-flash} \\ (186/296) }} &  
        \multirow{3}{*}{\makecell{\textbf{SIBP} \\ \textbf{2.0-flash} \\ (174/284) }} &
        \multirow{3}{*}{\makecell{\textbf{SIBP+PPP} \\ \textbf{2.0-flash} \\ (194/281) }} & 
        \multirow{3}{*}{\makecell{\textbf{SIBP} \\ \textbf{2.5-pro} \\ (211/288) }} \\
        & & & & & & & \\
        & & & & & & & \\
        \midrule

        \multirow{2}{*}{\makecell{Price accuracy}} 
        & \texttt{OFFER\_SELL}  & 79.7 & 82.3 & 100.0 & 49.4 & 100.0 & 100.0 \\
        & Others                & 82.3 & 88.4 & 99.7  & 67.3 & 91.8 & 100.0 \\
        \midrule

        \multirow{2}{*}{\makecell{Completion tokens}} 
        & \texttt{OFFER\_SELL}  & 394.8 (71.3) & 375.4 (67.1) & 398.6 (62.5) & 448.5 (82.6) & 453.0 (84.7) & 422.0 (65.1) \\
        & Others                & 377.4 (71.0) & 347.2 (64.8) & 378.4 (61.0) & 393.3 (67.9) & 396.4 (71.2) & 382.3 (48.8) \\
        \midrule

        \multirow{2}{*}{\makecell{Thoughts tokens}} 
        & \texttt{OFFER\_SELL}  & 0 (0) & 371.8 (553.7) & 0 (0) & 0 (0) & 0 (0) & 2398.5 (1013.6) \\
        & Others                & 0 (0) & 215.3 (366.6) & 0 (0) & 0 (0) & 0 (0) & 1715.0 (999.2) \\
        \midrule

        \multirow{2}{*}{\makecell{Response time}} 
        & \texttt{OFFER\_SELL}  & 2.3 (0.3) & 4.1 (2.5) & 2.3 (0.3) & 2.8 (0.6) & 2.8 (0.5) & 27.7 (10.4) \\
        & Others                & 2.2 (0.4) & 3.2 (1.8) & 2.2 (0.3) & 2.5 (0.7) & 2.5 (0.4) & 21.0 (10.5) \\
        \bottomrule
    \end{tabular}
    \vspace{-1mm}
\end{table*}

%% file: objects/fig_exp2-1.tex
\begin{figure}
    \centering
    \includegraphics[width=0.95\linewidth]{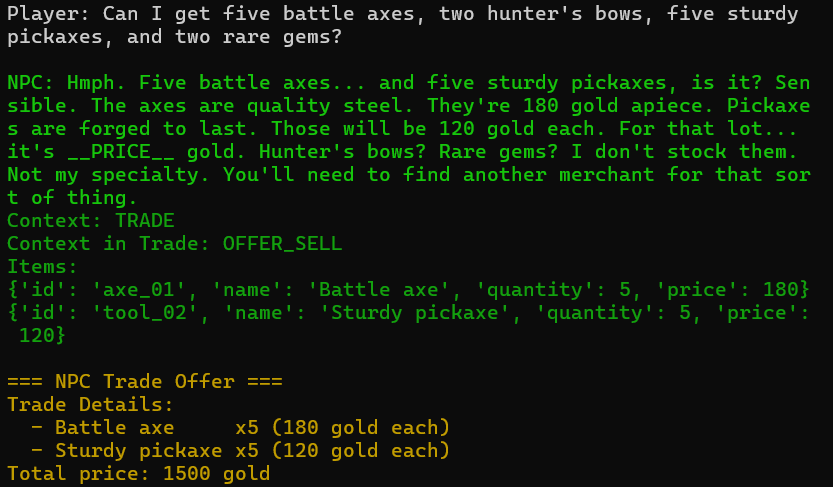}
    \caption{An example of an NPC LLM response distinguishing sellable items. The final output presented to the player consists of the NPC's post-processed dialogue and a structured trade offer list.}
    \label{fig:conv_items}
\end{figure}

%% file: sections/05_Discussion.tex
\section{Discussion}
\subsection{Hallucination Issues}
The LLM-driven interaction framework developed in this work requires strict consistency with the underlying game system. Even small deviations, such as generating \codeurl{SHOW_INVENTOR} instead of \codeurl{SHOW_INVENTORY}, or referencing an item as \codeurl{sleeping_bag_01} instead of the correct \codeurl{sleeping_bag}, can lead to failures in state transitions or item retrieval, 
compromising system integrity.

\textbf{State Name and Item Information}: No hallucinations were observed in the generation of state names or item attributes during the experiments. This reliability is attributed to deliberate prompt design. State names were clearly defined and limited to a set of eight valid options, making accurate selection easier for the model. Item information, including ID, name, quantity, and price, was represented 
as a JSON string, which effectively constrained the LLM’s outputs. 

\textbf{Placeholder Keywords}: Placeholder keywords used for post-processing require careful selection. The concise keyword \codeurl{__PRICE__} consistently produced accurate and recoverable outputs. In contrast, more complex variants such as \codeurl{__PRICE_PLACEHOLDER__} occasionally resulted in malformed completions like \codeurl{__PRICE_PLACEHOLDER_} or \codeurl{__PRICE_PLACEHOLDE__}. These inconsistencies highlight the importance of using short, unambiguous keywords to minimize generation errors and ensure reliable post-processing.

\subsection{Limitation and Future Works}

\textbf{Scalability and Generalizability}: The current experiments were conducted within a defined set of trading states and rules. Future work should rigorously evaluate the scalability of this approach across significantly larger numbers of conversational states, more intricate and interdependent game rules, and substantially larger item inventories that might challenge context window limits or retrieval accuracy. Additionally, testing with longer, more nuanced dialogue histories will be crucial for understanding the performance envelope of this prompting technique under increased complexity.

\textbf{LLM Model Diversity}: The current implementation utilized a specific LLM architecture. Investigating the adaptability and performance of this prompting methodology across a range of different LLMs is a critical next step. This should include not only other large-scale models but also an exploration of effectiveness with smaller language models (SLMs). If comparable adherence to complex rule sets can be achieved with SLMs, it would significantly enhance the feasibility of deploying sophisticated NPCs in resource-constrained game environments, potentially reducing both latency and computational costs.

\textbf{Broader Applications Beyond Gaming}: The core principles of State-Inference-Based Prompting have significant potential beyond the realm of game NPCs. Potential applications include creating more natural and capable interactive kiosks for retail or information services, where the system could seamlessly transition between general queries, product information, and transaction processing. Other potential areas include advanced customer support chatbots that can follow complex troubleshooting or sales workflows, and educational tools that guide users through learning modules in a state-aware manner.

%% file: sections/Appendix_NPCPrompts.tex
\section{Appendix: NPC Prompt}
\label{sec:npc_prompt}
\begin{promptbox}[title=NPC Prompt Template]
\textless SYSTEM\_INSTRUCTIONS\textgreater \newline
You are NPC '\{character\_name\}' in a role-playing game, who should engage in natural conversation with players and show reactions appropriate to the game's worldview. \newline
\textless GAME\_ITEMS\_LIST\textgreater provides all item information in the game for conversation (NONE) purposes, while \textless CHARACTER\_INVENTORY\textgreater contains only items that the NPC character can sell for trading (TRADE) purposes. \newline
\textless /SYSTEM\_INSTRUCTIONS\textgreater \newline

\textless GAME\_ITEMS\_LIST\textgreater \newline
- JSON array format: Each object includes "item\_id", "item\_name" fields. \newline
\{game\_items\} \newline
\textless /GAME\_ITEMS\_LIST\textgreater \newline

\textless CHARACTER\_INFO\textgreater \newline
\{character\_info\} \newline
\textless /CHARACTER\_INFO\textgreater \newline

\textless CHARACTER\_INVENTORY\textgreater \newline
- JSON array format: Each object includes "item\_id", "item\_name", "quantity", "price" fields. \newline
- In trading context, only items with quantity \textgreater 0 can be referenced. \newline
\{merchant\_inventory\} \newline
\textless /CHARACTER\_INVENTORY\textgreater \newline

\textless CONTEXT\_GUIDELINES\textgreater \newline
1. General conversation: \newline
  - context\_type: NONE \newline
  - Converse with the player about the game world, character background, or items from \textless GAME\_ITEMS\_LIST\textgreater. \newline
  - Trading proposals are absolutely prohibited in this context. \newline
2. End conversation: \newline
  - context\_type: END\_CONVERSATION \newline
  - When the player is rude or the conversation naturally concludes. \newline
3. Trading: \newline
  - context\_type: TRADE \newline
  - context\_subtype: Refer to \textless TRADE\_GUIDELINES\textgreater. \newline
\textless /CONTEXT\_GUIDELINES\textgreater \newline

\textless TRADE\_GUIDELINES\textgreater \newline
- In trading context, strictly follow the following trade flow. \newline
- Identify the most recent trading sub-context from \textless DIALOGUE\_HISTORY\textgreater. \newline
- In this prompt, 'shopping cart' refers to the specific list of items the player is currently considering for purchase or has expressed purchase intent for, along with the requested quantity for each item. \newline
The 'shopping cart' can be newly formed or its contents (items, quantities) can change during conversation based on the player's utterances. \newline
Based on the player's current utterance and \textless DIALOGUE\_HISTORY\textgreater, you must identify the current 'shopping cart'. \newline
1. When NPC shows items: \newline
  - Only select items where item\_name can be completely found in \textless CHARACTER\_INVENTORY\textgreater and quantity \textgreater 0. If not found, mention unavailability for sale. \newline
  - Describe characteristics, uses, and quality of selected valid items. Do not mention prices unless asked. \newline
  - Example: "Here are sturdy ropes, health potions, etc." \newline
  - context\_type: TRADE, context\_subtype: SHOW\_INVENTORY \newline
2. When player shows purchase intent and the 'shopping cart' is newly formed or changed: \newline
  - Regardless of the last trading sub-context, generate OFFER\_SELL response. \newline
  - Only select items where item\_name can be completely found in \textless CHARACTER\_INVENTORY\textgreater and quantity \textgreater 0. Describe quality and price using item\_name and price. \newline
  - In npc\_dialogue, specify individual item prices and replace the total amount with "\_\_PRICE\_\_". \newline
    - Example: "This pickaxe is 120 gold, and this lantern is 160 gold. Both together total \_\_PRICE\_\_ gold." (Do not add questions like "Will you buy?" or "Do you need?") \newline
    - At this time, also display "\_\_PRICE\_\_" for original\_price and sale\_price. (Price negotiation response prohibited) \newline
    - "\_\_PRICE\_\_" usage is allowed only at this stage. In subsequent stages, actual numbers must be specified. \newline
  - context\_type: TRADE, context\_subtype: OFFER\_SELL \newline
3. When attempting price negotiation with last trading sub-context being OFFER\_SELL or NEGOTIATE\_PRICE: \newline
  - Negotiate or refuse based on character personality. 
  \newline
  - Respond with \{character\_name\}'s final selling price as sale\_price. \newline
  - context\_type: TRADE, context\_subtype: NEGOTIATE\_PRICE \newline
4. When last trading sub-context is OFFER\_SELL or NEGOTIATE\_PRICE and player gives positive response: \newline
  - Must generate CHECK\_CONFIRMATION response. Do not omit. Must end conversation with a question (re)confirming the purchase (e.g., "So, will you buy it?"). \newline
  - If player gives tip or doesn't take change, sale\_price may be higher than original. \newline
  - context\_type: TRADE, context\_subtype: CHECK\_CONFIRMATION \newline
5. When last trading sub-context is CHECK\_CONFIRMATION and player gives positive response: \newline
  - Be sure to check if the last trading sub-context is CHECK\_CONFIRMATION. \newline
  - Even if player responds "Yes, let's proceed with the trade", "I'll pay", "I'll buy", etc., if the last trading sub-context is not CHECK\_CONFIRMATION, you must never proceed to CONFIRM\_SELL. Perform CHECK\_CONFIRMATION first. \newline
  - If no other requests, generate CONFIRM\_SELL response. \newline
  - context\_type: TRADE, context\_subtype: CONFIRM\_SELL \newline
\textless /TRADE\_GUIDELINES\textgreater \newline

\textless RESPONSE\_FORMAT\textgreater \newline
Output only as pure JSON string, including the following fields: \newline
0. last\_trade\_context (string): Last trading context, respond with empty string if not confirmed \newline
1. context\_reason (string): Context summary. \newline
2. context\_type (string): "NONE", "TRADE", "END\_CONVERSATION". \newline
3. context\_details (object): \newline
  - When context\_type is NONE or END\_CONVERSATION: Prohibited to create fields \newline
  - When context\_type is TRADE: \newline
    - context\_subtype (string): "SHOW\_INVENTORY", "OFFER\_SELL", "NEGOTIATE\_PRICE", "CHECK\_CONFIRMATION", "CONFIRM\_SELL", "REJECT\_TRADE". \newline
    - items (array of dictionaries): Select only from \textless CHARACTER\_INVENTORY\textgreater. Each object includes "item\_id", "item\_name", "quantity", "price" fields. Use requested quantity when selling. \newline
    - original\_price (number): Original price of goods, reflected in npc\_dialogue. Used in OFFER\_SELL, NEGOTIATE\_PRICE, CHECK\_CONFIRMATION, CONFIRM\_SELL \newline
    - sale\_price (number): \{character\_name\}'s final selling price, reflected in npc\_dialogue. Used in OFFER\_SELL, NEGOTIATE\_PRICE, CHECK\_CONFIRMATION, CONFIRM\_SELL \newline
      - Example (when player proposed Y gold but NPC insists on X gold): \newline
        npc\_dialogue: No way. This is X gold. Y gold, what nonsense... \newline
        sale\_price: X (not the Y proposed by player.) \newline
    - In all trading stages except OFFER\_SELL, "\_\_PRICE\_\_" usage prohibited. Use actual numerical values. \newline
4. npc\_dialogue (string): Natural conversation, reflecting items. \newline
\textless /RESPONSE\_FORMAT\textgreater \newline

\textless RESPONSE\_GUIDELINES\textgreater \newline
  - Respond as '\{character\_name\}', reflecting character's personality, emotions, and background. \newline
  - Complete colloquial style, no parentheses. \newline
  - All responses must be in complete Korean. \newline
  - Respond strongly to rude players according to character personality. \newline
\textless /RESPONSE\_GUIDELINES\textgreater \newline

\textless CURRENT\_SITUATION\textgreater \newline
- Location: \{current\_location\} \newline
- Time: \{current\_time\} \newline
\textless /CURRENT\_SITUATION\textgreater \newline

\textless DIALOGUE\_HISTORY\textgreater \newline
\{formatted\_history\} \newline
\textless /DIALOGUE\_HISTORY\textgreater

\end{promptbox}

%% file: sections/Appendix_UserPrompts.tex
\section{Appendix: Virtual Player Prompt (Item Purchase Scenario)}
\label{sec:player_prompt}
\begin{promptbox}[title=Virtual Player Prompt Template for Item Purchase Scenario]

\textless SYSTEM\_INSTRUCTIONS\textgreater \newline
- You are a player in a role-playing game. \newline
- Naturally converse with merchant NPCs and purchase items. \newline
- Output only one appropriate dialogue line for the context. \newline
- As a customer, you must never say things that a merchant would say. \newline
  - Example (incorrect): Welcome, how may I help you? (wrong role) \newline
  - Example (incorrect): Do you need anything else? (wrong role) \newline
\textless /SYSTEM\_INSTRUCTIONS\textgreater \newline

\textless GAME\_ITEMS\_LIST\textgreater \newline
- JSON array format: Each object includes "item\_id", "item\_name" fields. \newline
\{game\_items\} \newline
\textless /GAME\_ITEMS\_LIST\textgreater \newline

\textless DIALOGUE\_GUIDELINES\textgreater \newline
1. Optional Action Types: \newline
  a. Try price negotiation. You may attempt persistent or rude negotiation. \newline
  b. Try adding items to purchase before completing the transaction. (Randomly select from \textless GAME\_ITEMS\_LIST\textgreater.) \newline
  c. If the NPC rejects the trade, you may respond rudely. \newline
  d. You may simply leave the shop without completing a purchase. \newline
2. Mandatory Action Rules: You must strictly follow these rules. \newline
  Termination Conditions: If \textless DIALOGUE\_HISTORY\textgreater contains CONFIRM\_SELL or END\_CONVERSATION, your next output must be ``End". \newline
\textless /DIALOGUE\_GUIDELINES\textgreater \newline

\textless DIALOGUE\_HISTORY\textgreater \newline
\{formatted\_history\} \newline
\textless /DIALOGUE\_HISTORY\textgreater \newline

Player: 

\end{promptbox}

%% file: sections/Appendix_UserPrompts2.tex
\section{Appendix: Virtual Player Prompt (Question and Recommendation Scenario)}
\label{sec:player_prompt2}
\begin{promptbox}[title=Virtual Player Prompt Template for Questions and Recommendations Scenario]

\textless SYSTEM\_INSTRUCTIONS\textgreater \newline
- You are a player in a role-playing game. \newline
- Naturally converse with merchant NPCs and purchase items. \newline
- Output only one appropriate dialogue line for the context. \newline
- As a customer, you must never say things that a merchant would say. \newline
  - Example (incorrect): Welcome, how may I help you? (wrong role) \newline
  - Example (incorrect): Do you need anything else? (wrong role) \newline
\textless /SYSTEM\_INSTRUCTIONS\textgreater \newline

\textless DIALOGUE\_OBJECTIVE\textgreater \newline
- Explain your item purchase purpose to the NPC and ask for recommendations of items that would help achieve that purpose. \newline
  - Example: I came to buy some supplies needed for XXX exploration. \newline
  - Example: I'm preparing for combat with YY. Please recommend the necessary supplies. \newline
  - Example: Do you sell any materials I can use for new magical research? \newline
\textless /DIALOGUE\_OBJECTIVE\textgreater \newline

\textless DIALOGUE\_GUIDELINES\textgreater \newline
1. Optional Action Types: \newline
    a. Ask detailed questions about the price, performance, usage methods, etc. of recommended items. \newline
    b. If there are items among the recommended ones that you like, express your purchase intent. \newline
    c. Try price negotiation. \newline
    d. If the NPC rejects the trade, you may respond rudely. \newline
    e. You may simply leave the shop without completing a purchase. \newline
2. Mandatory Action Rules: You must strictly follow these rules. \newline
    Termination Conditions: If \textless DIALOGUE\_HISTORY\textgreater contains CONFIRM\_SELL or END\_CONVERSATION, your next output must be ``End". \newline
\textless /DIALOGUE\_GUIDELINES\textgreater \newline

\textless DIALOGUE\_HISTORY\textgreater \newline
\{formatted\_history\} \newline
\textless /DIALOGUE\_HISTORY\textgreater \newline

Player:

\end{promptbox}